# State-of-the-art Models for Object Detection in Various Fields of Application

Syed Ali John Naqvi
Dept of Computer Engineering
National University of Science and Technology
Lahore, Pakistan
syedalijohn@gmail.com

Syed Bazil Ali
Dept of Computer Engineering
National University of Science and Technology
Karachi, Pakistan
sbazil.ce41ceme@ce.ceme.edu.pk

*Abstract*— We present a list of datasets and their best models with the goal of advancing the state-of-the-art in object detection by placing the question of object recognition in the context of the two types of state-of-the-art methods: one-stage methods and two stage-methods. We provided an in-depth statistical analysis of the five top datasets in the light of recent developments in granulated Deep Learning models - COCO minival, COCO test, Pascal VOC 2007, ADE20K, and ImageNet. The datasets are handpicked after closely comparing them with the rest in terms of diversity, quality of data, minimal bias, labeling quality etc. More importantly, our work extends to provide the best combination of these datasets with the emerging models in the last two years. It lists the top models and their optimal use cases for each of the respective datasets. We have provided a comprehensive overview of a variety of both generic and specific object detection models, enlisting comparative results like inference time and average precision of box (AP) fixed at different Intersection Over Union (IoUs) and for different sized objects. The qualitative and quantitative analysis will allow experts to achieve new performance records using the best combination of datasets and models.

*Keywords—Object detection, Deep Learning, Computer Vision,*

## I. INTRODUCTION

In the field of computer vision, object detection has experienced a remarkable revolution and is now regarded as a challenging area of study. Finding one or more relevant objects from still images or videos is the prime objective of object detection. A wide range of significant methods, including Convolutional Neural Networks (CNN), Machine Learning (ML), Artificial Intelligence (AI), image processing, and pattern recognition are all part of it. Various computer vision applications for object detection are now commonly in practice. This includes surveillance and security, autonomous driving, anomaly detection, pedestrian detection, and robot vision [1].

Dataset and model are the core of elements of Machine Learning (ML) [2]. A dataset is a set of connected, distinct elements of connected data that may be viewed separately, together, or handled as a single unit [3]. A data structure of some kind is used to arrange a data set. The dataset, which falls under the purview of the object detection domain, consists of thousands of photos taken in a variety of lighting conditions, such as dawn, dusk, and several states in-between with respective bounding box coordinates and tags. [4]. After a dataset has been trained, a model is produced. Users may view the entire dataset structure through a model and then deal with dataset sections as per need. A file designed to detect specific patterns in dataset is known as a machine learning model. A model is trained using dataset and thorough an algorithm that enables it to interpret and understand from the data. They are crucial in making use of the dataset's content. No matter how balanced, relevant, and high-quality dataset you select, a poor model will still be unable to find a pattern in the training dataset, thus affecting the performance and accuracy of the machine learning model. And vice versa, a good model cannot make up for a poor dataset [5].

The performance of object detectors is getting better every day because of the rapid advancement in deep learning (DL) networks and GPU's computational capacity. However, there is argument amongst the experts regarding which model and approach is best to implement object detection in certain scenarios. Most specialists rate models according to their precision, speed, scene understanding of 2D and 3D objects and any other parameter of relevance [6]. However, there is a trade-off between these parameters in most computer vision applications.

The success of deep learning has drawn a lot of attention to object detection in recent years. However, there is gap in the knowledge observed related to the models introduced in the last couple of years in the object detection field. This paper attempts to fill the gap in research by discussing the object detection and classification datasets [7,8,9,10,11] along with the best suited models for specific fields.

There are a variety of datasets available, each with thousands of images, as training and testing for object detection. For instance, the ImageNet dataset has 14 million classified and tagged photos. Since every dataset is produced with a specific goal in mind, there is no single best dataset for every object detection problem. The target class should appear more frequently in a well-balanced dataset, for example, lung segmentation from Chest X-Ray dataset to diagnose lung abnormality in medicine [12]. The weapon detection dataset [13] in this case is an example of an unbalanced dataset. There are more minority class instances (no weapon) than majority class instances (weapon detected).

An in-depth discussion using qualitative and quantitative analysis allow experts to understand which combination of dataset and model will yield best results for their area of interest. It also discusses the difficulties in detecting objects in the past and how recent advancements in datasets and models have managed to address these issues.

## II. DISCUSSION

### A. COCO val2017 dataset

COCO val2017 was a change in the original COCO dataset of 2015 [14]. The greatest difference to the COCO val2017 dataset in 2017 was the modification of the split from 83K/41K train/val to 118K/5K for train/val in response to user suggestions. Also, there are no additional labels for detection or descriptors, and the original images database is used. Interestingly, 40K train images and 5K val images both had stuff annotations appended to them. Images from COCO that are unlabeled but share the same class labels as the labelled images might be helpful for COCO's semi-supervised learning. With the use of a significant quantity of unlabelled data, semi-supervised object detection (SSOD) attempts to simplify the training and implementation of object detectors [15].

*1) On COCO val2017, DINO-Swin model of 2022 produces the best performance. Using multi-scale characteristics, DINO surpasses all existing ResNet-50-based models on COCO val2017 in both the 12-epoch and the 36-epoch configurations]. It shows an average precision Mean Average Precision (mAP) of 63.2 [16].*

*2) DINO-s5cale achieves 49.4 AP in 12 epochs and 51.2 AP in 24 epochs using the ResNet-50 backbone - a convolutional neural network with 50 layers. More than a million images from the COCO val2017 database may be employed to train it. The trained network is able to categorize photos into 1000 different item categories, including several different animals, tech accessories and stationary items. The 4-scale model produces comparable results.*

*3) The Swin Transformer V2 model, the biggest dense vision model to date, allows for training with pictures with a resolution of more than 15 billion. The models provide 62.5 AP-box and 53.7 AP-mask. On four exemplary computer vision applications, including ImageNet-V2 image classification, COCO object recognition, and ADE20K semantic segmentation, it achieved historic performance records [17].*

*4) Florence achieves state-of-the-art performance in its 2022 Florence CoSwin model for object detection, including 62.4 mAP on the COCO val2017 data set [18].*

*5) The Vision Transformer (ViT) model, which employs the pure transformer concept effectively applied to series of image regions, outperforms in image categorization tasks. When compared to state-of-the-art convolution networks, Vision Transformer (ViT) achieves great results with far less computational overhead during training [19]. With only ImageNet-1K pre-training, it can get up to 61.3 APbox on the COCO dataset and 53.1 APmask, competing with the prior state-of-the-art models for object detection that were all built on hierarchical backbones [20].*

TABLE I.

| Model Year | COCO val2017 dataset | | |
|---|---|---|---|
| | *Model name* | *AP-box* | *AP-mask* |
| 2022 | Dino-Swin | 63.1 | - |
| 2022 | Dino 5-scale | 51.2 | - |
| 2022 | Swin V2-G | 62.5 | 53.7 |
| 2022 | Florence CoSwin | 62.4 | - |
| 2022 | ViT-H | 61.3 | 53.1 |

Fig. 1. Best models for COCO val2017 in 2022.

### B. COCO test dataset

It features approximately 1.5 million object instances and image annotations in 80 categories for a narrow collection of the items we come across in our everyday life. [21]. State-of-the-art neural networks are mostly used in the demanding, high-quality visual datasets included in the COCO dataset for computer vision. COCO is frequently used to assess the performance of real-time object detection algorithms, including face detection, video streams, and weapon surveillance.

*1) DINO-Swin is an unlabeled, self-distillation learning model that accurately predicts the results. The model dramatically decreases the size of the pre-training data while still outperforming other models on the list. It produces (mAP) of 63.3 with COCO test dataset [16].*

*2) According to a study [22], feature distillation (FD), a straightforward post-processing technique, may dramatically enhance the poor fine-tuning performance of pre-training techniques. A unique model was required, and the SwinV2-G model enhanced the fine-tuning accuracy on COCO object identification by +1.5 mIoU / +1.1 mAP to 61.4 mIoU / 64.2 mAP, setting new standards of performance.*

*3) Florence CoSwin-H performs well with the large majority of 44 representative benchmarks Florence has exceptional performance in a variety of transfer learning techniques, including few-shot transfer and zero-shot transfer for unexpected pictures and items. It gives 62.4 AP-box with COCO test dataset [18].*

*4) When combined with the COCO test dataset, BEiT-3 delivers state-of-the-art performance on both vision and vision-language tasks. On pictures, words, and their combinations, BEiT-3 conducts masked modelling in an integrated manner [23].*

*5) Swin V2-G uses 40 times less labelled data and 40 times less training time than Google's billion-level visual models, proving that its training is far more effective [17].*

TABLE II.

| Model Year | COCO test dataset | | |
|---|---|---|---|
| | *Model name* | *AP-box* | *AP-mask* |
| 2022 | Dino-Swin | 63.3 | - |
| 2022 | FD-SwinV2-G | 64.2 | 55.4 |
| 2022 | Florence CoSwin-H | 62.4 | - |
| 2022 | BeiT-3 | 63.7 | 54.8 |
| 2022 | Swin V2-G | 63.1 | 54.4 |

Fig. 2. Best models for COCO test in 2022.

### C. Pascal VOC 2007

There are 20 object categories in Pascal VOC 2007 dataset, including:

- People
- Vehicles include bicycles, boats, buses, cars, motorcycles, and trains.
- Bottle, chair, dining room table, potted plant, sofa, television/monitor in indoors.
- Bird, cat, cow, dog, horse, sheep, and other animals

The training dataset is a collection of photos, each of which contains an annotation file including a bounding box and an object class label for all twenty classes represented in the image. The dataset may be useful for applications including object counting and multilevel classification.

*1) When combined with the Pascal VOC 2007 dataset, Florence effectively expands to many tasks across space, time, and modal with excellent transferability, and delivered record performance on a number of visual benchmarks [18]. Applications like depth/flow estimates, tracking, and other vision+language tasks can benefit most from the combination.*

*2) A significant difficulty in computer vision is creating instance segmentation models that are data-efficient and can manage uncommon object types. A possible approach to solving this problem is to use data augmentations, such as Copy-Paste augmentation. The model gives 76.5 AP-box performance [24].*

TABLE III.

| Model Year | Pascal VOC 2007 dataset | |
|---|---|---|
| | *Model name* | *AP-box* |
| 2022 | Florence CoSwin-H | 90.5 |
| 2021 | Copy-Paste | 76.5 |

Fig. 3. Best models for Pascal VOC 2007 dataset in 2022.

### D. ADE20K

More than 20K scene-centric photos with detailed labels for objects and object sections are included in the ADE20K dataset. There are 150 semantic categories in all, which cover elements like grass, sky, and roads as well as distinct items like people, cars, and beds.

The dataset is useful for applications including semantic segmentation and image-to-image translation.

*1) It is pre-trained on enormous monomodal and multimodal data that gives 62.8 AP-box with ADE20K dataset for object detection application [23].*

*2) HorNet-L performs much better than Swin Transformers and ConvNeXt with comparable overall design and training parameters [25]. HorNet also exhibits good scalability to bigger model sizes and additional training data.*

*3) When it came to solving the quadratic computational inefficiency issue, Focal-L (Dyhead) performed exceptionally well, especially for high-resolution vision tasks like object detection. For semantic segmentation, it offered 55.4 mIoU on ADE20K, making new SoTA on three of the most difficult computer vision challenges [26].*

TABLE IV.

| Model Year | ADE20K dataset | |
|---|---|---|
| | *Model name* | *mIOU* |
| 2022 | BeiT-3 | 62.8 |
| 2022 | Hornet-L | 57.9 |
| 2022 | Focal-L (Dyhead) | 55.4 |

Fig. 4. Best models for ADE20K dataset in 2022. (*figure caption*)

### E. ImageNet

The ImageNet dataset is a massive collection of human photos with annotations generated by an academic project for the creation of computer vision algorithms. The work has manually annotated more than 14 million photographs to identify the things they portray, and more than one million of the images also include bounding boxes [27]. Researchers now have a shared collection of images to compare their models and algorithms against ImageNet.

The dataset has a useful use case in neural architecture search and self-supervised image classification.

*1) SwinV2-G large-scale computer vision model that successfully addresses the three key challenges of training inconsistency, resolution gaps between pre-training and fine-tuning, and reliance on labelled data [17].*

*2) As discussed earlier, Focal-L is a great model to address the quadratic computational inefficiency issues.*

*3) For visual processing, the Vision GNN (ViG) architecture is an an excellent that works on the principle of extracting graph-level features. Each image is divided into a number of portions, which are considered as nodes and are connected to one another to form a graph [28]. ViG model is made to convert and communicate information across all the nodes based on that graph.*

*4) MaxViT works at the state-of-the-art in a variety of conditions. For example, MaxViT obtains 86.5% ImageNet-1K top-1 accuracy without additional data, whereas our model gets 88.7% top-1 accuracy with ImageNet-21K pre-training [29]. MaxViT as a foundation provides good performance on object detection as well as visual aesthetic evaluation for downstream operations. Also, the model has great dynamic modelling capabilities [30].*

*5) BeiT-3 is an state-of-the-art model for object detection used commonly with multiway transformer network, which the experts employs as their backbone for encoding various modalities [23].*

TABLE V.

| Model Year | ImageNet dataset | |
|---|---|---|
| | *Model name* | *ImageNet-1K Top-1 accurracy* |
| 2022 | SwinV2-G | 90.2 |
| 2021 | Focal-L | 83.8 |
| 2022 | ViG-S | 82.1 |
| 2022 | MaxVit-B | 86.7 |
| 2022 | BeiT-3 | 89.6 |

Fig. 5. Best models for ImageNet dataset in 2022.

## III. CONCLUSION

In this work, we intend to capture the major developments made in last couple of years using state-of-the-art models to achieve excellent outcomes in the object detection domain.
One of the most prevalent task categories in computer vision is object detection, and it is now becoming increasingly common in a variety of real-life scenarios. We made an effort to offer the most effective fusion of object detection datasets with newly invented models. The best models and their ideal use cases are listed in the article to allow researchers and experts to achieve record performance of their overall model. The paper offers a thorough analysis of several general and object detection models, including comparisons of inference time and average accuracy of boxes (AP). This qualitative and quantitative analysis is a great blueprint to help anyone pick the best dataset and model combination for their object detection project.